\documentclass[10pt,twocolumn,letterpaper]{article}

\newcommand{\papermode}{2} %
\ifcase\papermode
    \usepackage[review]{cvpr}       %
\or
    \usepackage{cvpr}               %
\or
    \usepackage[pagenumbers]{cvpr}  %
\fi

\usepackage{amsmath}

\usepackage{overpic}
\usepackage{enumitem} %
\usepackage{overpic} %
\usepackage{color}
\usepackage{colortbl} %
\usepackage{adjustbox} %
\usepackage{tabularx} %
\usepackage{soul} %
\usepackage{dblfloatfix}
\usepackage{wrapfig}
\usepackage[T1]{fontenc}
\usepackage{lipsum} 
\usepackage{graphicx}
\usepackage{float}
\usepackage{placeins}

\definecolor{teal}{rgb}{0,.5,.5}
\definecolor{ligh_teal}{rgb}{0.51,.79,.77}

\definecolor{blue}{rgb}{0,0,1}
\definecolor{red}{rgb}{1,0,0}
\definecolor{green}{rgb}{0,1.,0}
\definecolor{orange}{rgb}{0.75, 0.4, 0}
\definecolor{turquoise}{cmyk}{0.65,0,0.1,0.3}
\definecolor{purple}{rgb}{0.65,0,0.65}
\definecolor{darkjunglegreen}{rgb}{0.1, 0.14, 0.13}
\definecolor{darkgreen}{rgb}{0.0, 0.2, 0.13}
\definecolor{darkred}{rgb}{0.6, 0.1, 0.05}
\definecolor{blueish}{rgb}{0.0, 0.3, .6}
\definecolor{light_gray}{rgb}{0.7, 0.7, .7}
\definecolor{pink}{rgb}{1, 0, 1}
\definecolor{greyblue}{rgb}{0.25, 0.25, 1}
\definecolor{magenta}{rgb}{1., 0., 1.}

\newcommand{\xgcc}[1]{{}}

\usepackage{blindtext}

\renewcommand{\paragraph}[1]{\vspace{1em}\noindent\textbf{#1}.}

\definecolor{cvprblue}{rgb}{0.21,0.49,0.74}
\usepackage{hyperref}
\hypersetup{pagebackref,breaklinks,colorlinks,citecolor=cvprblue}

\usepackage[capitalize]{cleveref}

\def\paperID{*****} %
\def\confName{X\xspace}
\def\confYear{X\xspace}

\title{An Object is Worth 64x64 Pixels: \\
Generating 3D Object via Image Diffusion}

\author{
Xingguang Yan$^{1}$\hspace{.4cm}
Han-Hung Lee$^{1}$\hspace{.4cm}
Ziyu Wan$^{2}$\hspace{.4cm}
Angel X. Chang$^{1,3}$
\\
$^{1}$Simon Fraser University\hspace{.3cm}
$^{2}$City University of Hong Kong\hspace{.3cm}
$^{3}$Canada-CIFAR AI Chair, Amii
\\
\\
\texttt{\href{https://omages.github.io/}{omages.github.io}}
}
\begin{document}
\twocolumn[{%
\renewcommand\twocolumn[1][]{#1}%
\maketitle

\vspace{-10pt}
\includegraphics[trim={0 0px 0 0},clip,width=\linewidth]{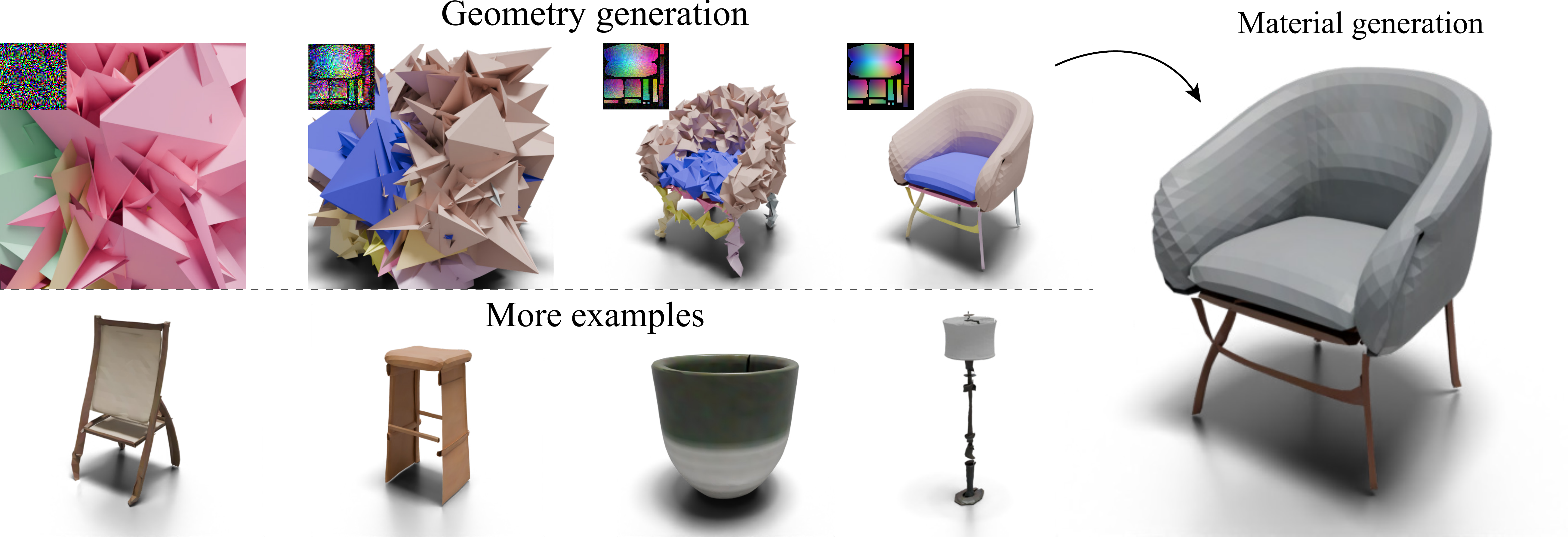}
\vspace{-20pt}
\captionof{figure}{Visualization of geometry generation (top row) using diffusion for Object Images followed by material generation (right). The spatial coordinates (xyz) are visualized as rgb colors (see inset Object Images). The colors of the denoising mesh highlight different connected components. After generating the geometry, our model can generate PBR materials given the geometry as a condition. Other examples of generated shapes are shown in the 2nd row. %
}
\vspace{10pt}
\label{fig:teaser}

}]

\begin{abstract}
\vspace{-8pt}
We introduce a new approach for generating realistic 3D models with UV maps through a representation termed "Object Images." This approach encapsulates surface geometry, appearance, and patch structures within a 64x64 pixel image, effectively converting complex 3D shapes into a more manageable 2D format.
By doing so, we address the challenges of both geometric and semantic irregularity inherent in polygonal meshes. This method allows us to use image generation models, such as Diffusion Transformers, directly for 3D shape generation. Evaluated on the ABO dataset, our generated shapes with patch structures achieve point cloud FID comparable to recent 3D generative models, while naturally supporting PBR material generation.

\end{abstract}

\vspace{-10pt}

\section{Introduction}
\begin{figure*}[t]
\centering
\includegraphics[width=1.0\linewidth]{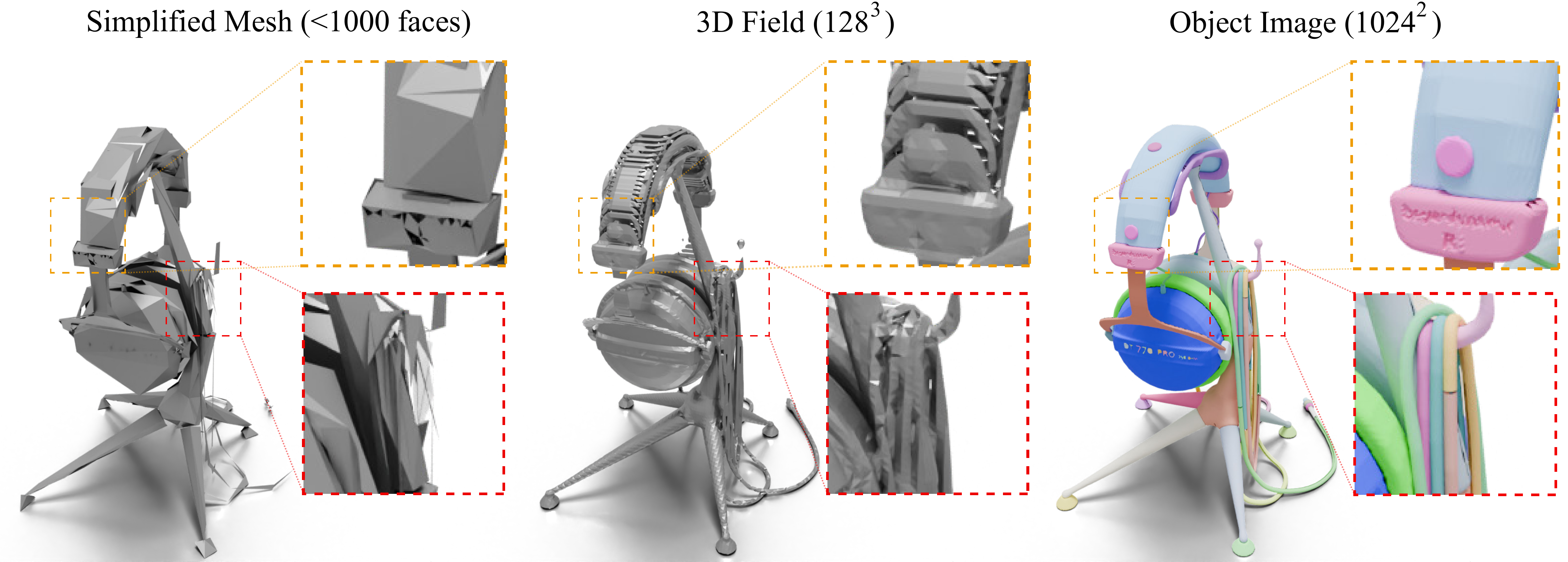}
\vspace{-2.0em}
\caption{Comparison of different representations used for generation. Simplified meshes (left) often introduce topological errors and degenerated parts. Volumetric representations (middle) tend to merge touching parts together, struggle to model thin surfaces, and cannot handle open surfaces. 
In contrast, Our Object Images (right) effectively preserve the topology and structure of the original mesh. %
}
\label{fig:repr_cmp}
\end{figure*}

\label{sec:intro}
Modeling high-quality 3D shapes is vital for industries such as film, interactive entertainment, manufacturing, and robotics. However, the process can be arduous and challenging. Inspired by the success of image generation models, which have significantly enhanced the productivity of 2D content creators~\cite{rombach2021highresolution}, researchers are now developing generative models for 3D shapes to streamline the synthesis of 3D assets~\cite{lee2024text,li2024advances}. 
Two challenges of building generative models for 3D assets are \textbf{geometric irregularity} and \textbf{semantic irregularity}.
\textbf{First}, unlike 2D images, standard 3D shape representations, such as polygonal meshes, are often highly \textit{irregular}; their vertices and connectivity do not follow a uniform grid and vary significantly in density and arrangement. Moreover, these shapes often \textit{possess complex topologies}, characterized by holes and multiple connected components, making it challenging to process meshes in a standardized way. These complexities pose a significant hurdle in generative modeling, as most existing techniques are designed for regular, tensorial data input. 
\textbf{Second}, 3D assets often possess rich \textit{semantic sub-structures}, such as parts, patches, segmentation and so on. These are not only essential for editing, interaction, and animation, but also vital for shape understanding and 3D reasoning.
However, these sub-structures also vary greatly, further hindering the design of generative models.

Most prior approaches only tried to handle either \textit{geometric irregularity} or \textit{semantic irregularity}~\cite{sid20203dstructgen}, but not at the same time. Many works bypass the former by converting original 3D shapes into more regular representations such as point clouds~\cite{zeng2022lion,nichol2022pointe}, implicit fields~\cite{zhang20233dshape2vecset,zhang2024clay}, or multi-view images~\cite{tang2024mvdiffusion++,xu2024instantmesh,siddiqui2024meta3dassetgentexttomesh}.
Although these formats are easier to process with neural networks, the conversions, both in forward and reverse directions, discard both geometric and semantic structures. The loss in information can significantly impact the representation accuracy and utility of the generated 3D models in applications. For example, for the headset in \cref{fig:repr_cmp}, implicit conversion fuses all the cables together, making the model difficult to use in animation.
Researchers have also tried to directly model geometric irregularity~\cite{nash2020polygen,siddiqui2023meshgpt, alliegro2023polydiff}, but are often restricted to simple meshes with less than 800 faces.

In our work, we explore to address the two irregularities \textit{simultaneously} by generating 3D shapes as Multi-Chart Geometry Images (MCGIM)~\cite{sander2003multi}, see~\cref{fig:teaser}.
Proposed over 20 years ago, Geometry Images~\cite{gu2002geometry,sander2003multi} addresses the geometric irregularity of meshes by decomposing the shape surface into one or multiple 2D patches that can be mapped and packed in a regular image. Through the irregular 2D shape packing process, MCGIM also efficiently addresses the semantic irregularity, that is, storing shapes of arbitrary number of patches in a single fixed-size image.
However, its automatic patch decomposition often results in less semantically meaningful patches and boundaries.
Our key observation is that many human-modeled 3D assets come with a semantically rich decomposition of patches in the form of UV-charts, which is usually only used for texturing in prior arts~\cite{Zeng2023Paint3DPA}.
These UV-charts can be easily processed to MCGIMs that can then be mapped back to 3D shapes.

Thus, inspired by Geometry Images, we propose to rasterize the mesh geometry (together with texture and material maps) into a 12-channel image as a new representation for 3D generation. This approach allows us to represent 3D shapes as 2D images and provides several benefits.  The representation 1) is simple and regular, 2) preserves the geometric and semantic structure together with PBR materials and 3) can be learned with image-based generative model to generate textured 3D meshes. We use the term \emph{Object Images} (\emph{omages} for short) 
for this representation, emphasizing its ability to encapsulate not just the geometry structure, but also material and semantically meaningful patch-decomposition of an object, and highlighting its potential for 3D generation by leveraging existing image-based methods. In this work, we convert the shapes of the ABO dataset ~\cite{collins2022abo}, which contains triangle meshes with designer-made UV-maps, into $1024$ resolution omages, downsample them to $64$ resolution with special care and use Diffusion Transformers \cite{Peebles2022DiT} to model their distribution. Our results show that our method generate shapes with patch structures that approaches similar geometric quality as state-of-the-art 3D generative models (in terms of point cloud FID), while naturally supporting PBR material generation.

\xgcc{
Logic flow:
Categorized by representations: Its properties, and the generative models built on it.
Polygonal meshes: 
    Characteristics: irregular, efficient. 
    Earlier works based nn: GeoCNN, MeshCNN, ...
    Non-native mesh generation: SDM-Net, AtlasNet, BSP-Net
    Near-native: polygen, meshgpt, polydiff: need decimation, erroneous
    Compare: decimation->erroneous, non-native for texture
Multi-chart: AtlasNet, 
    Characteristics: 
    GIM, MCGIM, IsoCharts, 
    PointUVDiffusion, Paint3D
    AtlasNet, atlasnet++, 
3D Field based representations
    neural implicit field: 
        SF, ASDF, Shape-E, S2VS, MSDF, DMTet
        Get3D, CC3D, 
        DreamFusion, Magic123, CAD, 
    Multi-view, videos
        Wonder3D, MVD++, ...
        slice3D

}

\section{Related Work}
\label{sec:related}

Our work lies in the field of surface shape generation. In this section, we present a survey of representative approaches categorized by their underlying 3D representation, with a focus on generative modeling.

\paragraph{Polygonal meshes}
As the most ubiquitous 3D representation, meshes, especially those modeled by 3D designers, are efficient and flexible, but also are well known for their difficulty to process with neural networks due to their irregularity. 
While various convolutional neural networks have been developed for mesh data~\cite{masci2015geodesic,poulenard2018multidirectional,hanocka2019meshcnn,sharp2021diffusion}, they have predominantly focused on shape understanding tasks like classification. The complexity of developing context-free unpooling operator on meshes impedes their use for mesh generation. To avoid the challenges of directly learning meshes with their native connectivity, researchers approximate the geometry using various surrogate mesh representations like surface patches~\cite{groueix2018atlasnet}, predicted meshes~\cite{dai2019scan2mesh}, deformed cuboids~\cite{gao2019sdmnet}, and binary space partitions~\cite{chen2020bsp}. 
This comes with the price of losing the details and structures of the original mesh.
In contrast, PolyGen~\cite{nash2020polygen} directly learns the distribution of the native mesh in a vertices-then-face manner with two autoregressive transformers~\cite{vaswani2017transformer}. However, this complex two stages pipeline exhibits limited robustness during inference as described in a later work, MeshGPT~\cite{siddiqui2023meshgpt}, which avoids this complexity by first encoding meshes into sequences of graph neural networks encoded face tokens that can be easily processed with a single autoregressive transformer.  MeshAnything~\cite{chen2024meshanything} further improves MeshGPT's encoder/decoder and enables conditional mesh generation given a reference point cloud.
These remarkable breakthroughs enable mesh generation with up to 800 triangular faces.
However, high-quality human-designed meshes usually have many more faces. For example, in the ABO dataset~\cite{collins2022abo}, over $70\%$ of the shapes has more than $10^4$ triangles. 
The current approaches need to first decimate the meshes to less than 800 faces, which may introduce topological errors (see~\cref{fig:repr_cmp}).
Moreover, these meshes often are accompanied with PBR materials and patch structures that polygonal mesh can not natively represent. In contrast, our object image is not restricted by the number of faces and naturally encapsulates material and patch information.

\begin{figure*}[t]
\centering
\includegraphics[trim={0 10px 0 8px},clip,width=1.0\linewidth]{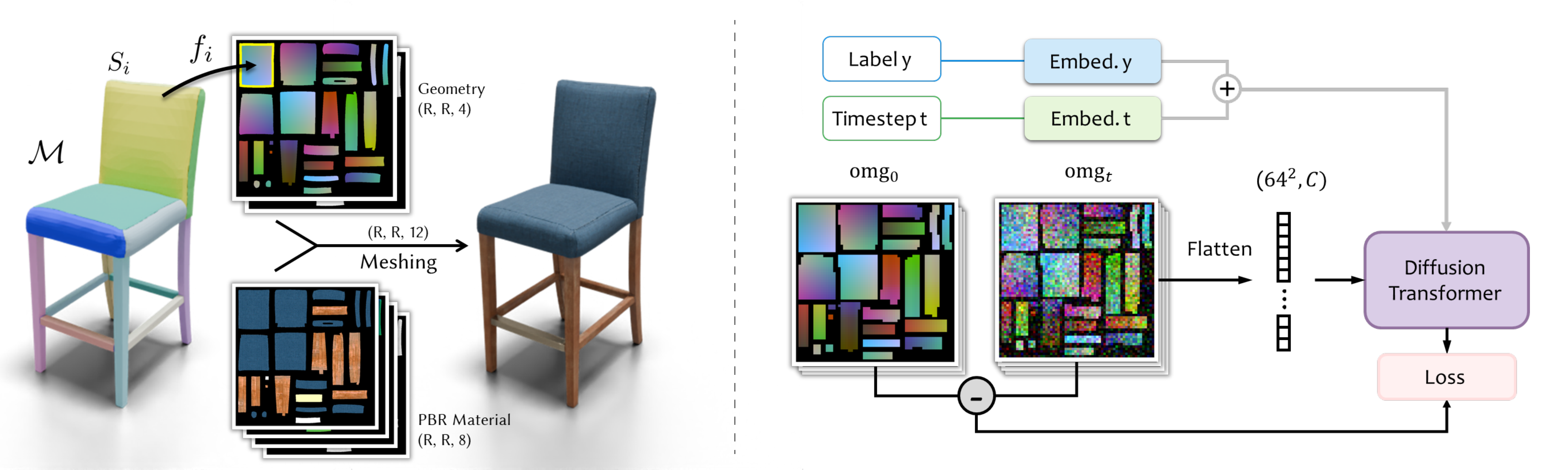}
\vspace{-15pt}
\caption{Method overview. \textbf{Left}: We assume the mesh $\mathcal{M}$ has patch decomposition $\{S_i\}$, and has single-valued uv-map $f_i$ that flattens patch $S_i$ into the 2D uv-domain. Together with the material maps, Object Images can represent high-quality photo-realistic object. \textbf{Right}: We train the image diffusion generative model with Diffusion Transformer. The input noised Object Image, omg, is first flattened into a sequence before passing into the transformer to predict the clean $\text{omg}_0$.}
\label{fig:method}
\end{figure*}

\paragraph{Multi-chart representations}
Modeling 3D shapes as single or multiple parametric patches (charts), is a prevalent approach for modeling smooth, curved shapes~\cite{goldman2009integrated,Piegl1995TheNB}.
Representing a polygonal mesh with parametric patches, commonly referred to as UV-mapping, aligns the irregular mesh with a regular 2D plane.
This alignment is essential for texture mapping, which paints rich textural images onto the 3D geometry~\cite{Catmull1974ASA,Blinn1976TextureAR}, and for surface remeshing, editing, and many other applications~\cite{sheffer2007meshparasurvey}.
To represent and store the geometry in a fully regular manner, Geometry Images~\cite{gu2002geometry,Hoppe2004OverviewOR} first parameterizes a mesh onto a planar domain and resamples the geometry onto an image pixel grid.
Further, Multi-charts Geometry Images (MCGIM) \cite{sander2003multi,zhou2004iso,carr2006rectangular} proposed to pack multiple patches into a single image, achieving lower distortion and is applicable to shapes with arbitrary topology.
Our proposed \textit{Object Image} is a kind of MCGIM extended with materials built specifically for image diffusion models.

While the utility of geometry images in deep learning has been well recognized, their use has been limited to either simple topologies or with automated patch splitting, making it challenging to obtain good surface parameterizations. \citet{Sinha2016DeepL3} and \citet{Maron2017ConvolutionalNN} applied CNNs to geometry images representing parameterizations on spherical and toric domains, respectively. 
Later, \citet{ben2018multi} and \citet{Alhaija2022XDGANM3}~(XDGAN) used GANs to generate \textit{genus-zero} shapes as geometry images.
Meanwhile, FoldingNet~\cite{yang2017foldingnet}, AtlasNet~\cite{groueix2018atlasnet} and its followups~\cite{Deprelle2019atlasnetv2, Williams2018DeepGP, Bednark2019DSP, Deng2020BetterDSP, Lei2020Pix2SurfLP, Deprelle2022LearningJS} have explored learning to approximate shapes with parametric patches in an unsupervised manner.
These efforts commonly employ algorithmic or approximated patch splitting, which tends to be either topologically constrained or inaccurate.

In contrast, we recognize that human-authored UV-atlases can be easily processed into MCGIMs, supporting arbitrary patch topology, and can be easily generated with image diffusion models.
While UV-atlases are widely used in recent learning-based mesh texturing methods~\cite{Zeng2023Paint3DPA}, they serve primarily as auxiliary information. In contrast, we note that UV-atlases effectively transform a mesh into parametric surfaces, providing a valuable representation for both geometry and material generation.
More recently, BrepGen~\cite{xu2024brepgen} synthesizes CAD B-Rep models by generating their patches and edges with diffusion models. However, it is still restricted to simple genus-zero patches and can only be applied to B-Rep models. 

\paragraph{3D fields and multi-view images}
3D shapes can be implicitly represented as a level-set of a spatial field. In this way, the irregularity challenge is circumvented, although important structural and topological information is inevitably gone along the way (See~\cref{fig:repr_cmp}).
Instead of generating a field directly as a 3D grid (voxels)~\cite{wu20163dgan}, the recent trend is to first parameterize the field with a neural network (neural field). 
Seminal works utilize auto-encoders~\cite{zhiqin2019imnet, lars2019occnet} or auto-decoders~\cite{park2019deepsdf} to compress the neural field as a single latent vector, which can be easily generated via methods like GANs~\cite{goodfellow2014generative} or VAEs~\cite{Kingma2013vae}. 
Later works represent and generate neural fields using multiple latent vectors to enhance spatial reasoning~\cite{Ibing20213DSG, Zheng2022SDFStyleGANIS, Mittal2022AutoSDF, Cheng2022SDFusionM3, Chou2022DiffusionSDFCG, Zheng2023LASDiff, Gao2022GET3DAG}. %
In particular, ShapeFormer~\cite{yan2022shapeformer}, 3DILG~\cite{zhang20223dilg}, 3DShape2VecSet~\cite{zhang20233dshape2vecset} and Mosaic-SDF~\cite{yariv2023mosaic} utilize the sparsity of the 3D shape to further compress the field and enables generating higher-resolution results.

Another line of work represents and generates shapes as multi-view images~\cite{tang2023mvdiffusion, tang2024mvdiffusion++, long2023wonder3d, liu2023syncdreamer, xu2024instantmesh}.
They adopt diffusion models to generate multiple 3D consistent images of different views. Meshes can then be reconstructed via neural field methods like NeRF~\cite{mildenhall2021nerf} or NeuS~\cite{Wang2021NeuSLN}. To enhance understanding of the shape structure, especially the interior, Slice3D~\cite{wang2023slice3d} proposes using images of shape slices instead.
Since most of these multi-image methods obtain geometry through neural fields, they share similar advantages and disadvantages with 3D field generation methods.

Our object image can be seen as a combination of neural field and mesh representations. It preserves the topology and patch structure of the original mesh while functioning as a specialized form of 2D neural field, making them highly suitable for neural network processing due to their regular structure.

\section{Method}
\label{sec:method}
In this section, we first present the mathematical formulation of the Object Image (\textit{omage} for short) representation (\Cref{sec:problem}). Next, we describe how to utilize image-based generative models, specifically Diffusion Transfomer~\cite{Peebles2022DiT}, to generate these omages (\Cref{sec:method-generation}). Finally, we describe how to obtain omages from a 3D asset (\Cref{sec:method-object-images}).

\begin{figure*}[t]
\centering
\includegraphics[trim={0 10px 0 8px},clip,width=1.0\linewidth]{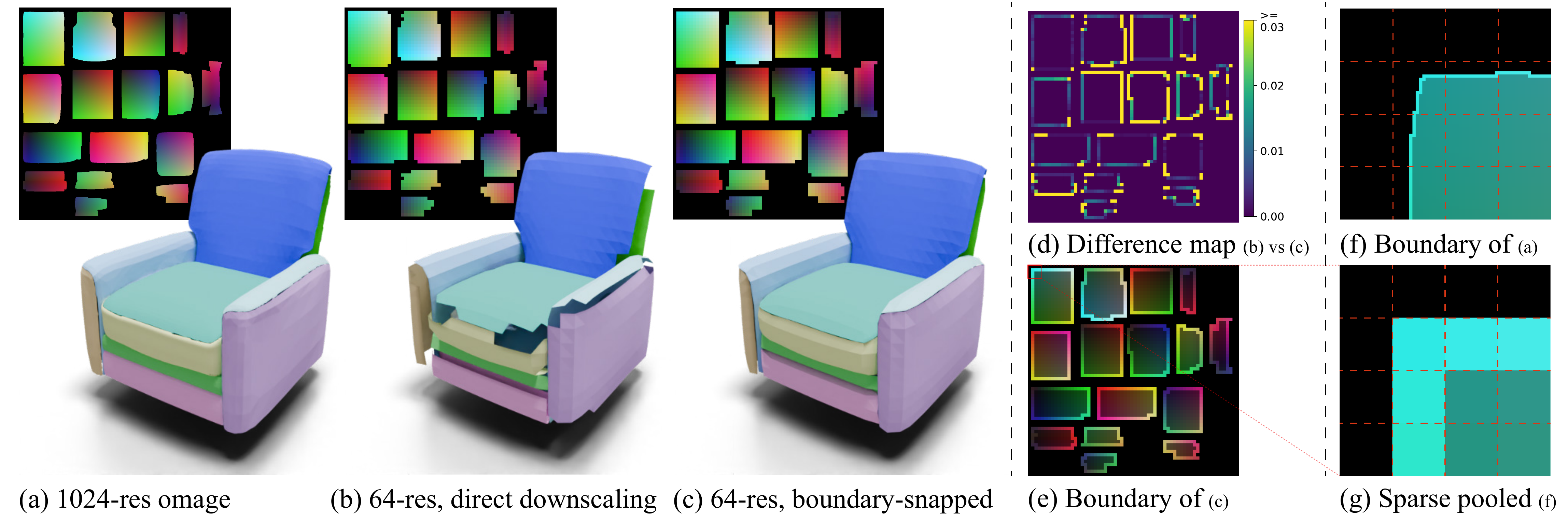}
\vspace{-15pt}
\caption{
Direct downscaling an omage from high-resolution (a) to lower resolution (b) usually leads to significant gaps between patches. By snapping the boundary vertices of the high resolution omage (f) into lower resolution via sparse pooling (e)(g), the gaps are significantly reduced (c)(d). 
}
\label{fig:edgesnapping}
\end{figure*}

\subsection{Object Images}
\label{sec:problem}

Given a 3D shape \( \mathcal{M} \) that is a 2D manifold embedded in 3D space, we consider it as a disjoint union of a set of \( N \) surface patches \( \{S_i\} \). By disjoint, we mean any two distinct patches \( S_i \) and \( S_j \) only overlap on their boundaries, i.e., \( S_i \cap S_j = \partial S_i \cap \partial S_j \). We assume each \( S_i \) is assigned a injective \emph{UV-mapping}, \( f_i: S_i \to [0,1]^2 \), where \( f_i(p) = (u, v) \) and \( [0,1]^2 \) is the \emph{UV-space}. The domain and image of \( f_i \) together are called an \emph{UV-island}, or \emph{UV-chart}: \( I_i = (S_i, f_i(S_i)) \). We denote the set of the \( N \) UV-islands, \( I := \{I_i\} \), as the \emph{UV-atlas} of \( \mathcal{M} \).

We indicate if a point in UV-space is inside an island by defining an occupancy function \( \alpha \) as follows:
\[
\alpha(u, v) = 
\begin{cases} 
1 & \text{if } \exists\, i \text{ such that } (u,v) \in f_i(S_i) \\
0 & \text{otherwise}
\end{cases}
\]

We then define the position map \( \pi \) of \( M \) as follows:
\[
\pi(u, v) = \begin{cases} f_i^{-1}(u, v) & \text{if } (u,v) \in f_i(S_i) \text{ for some } i \\ \text{undefined} & \text{otherwise} \end{cases} 
\]

By \emph{packing} the UV-islands through translation and scaling, we ensure that the set family \( \{f_i(S_i)\} \) is disjoint, making \( \pi \) and \( \alpha \) strictly deterministic (single-valued). Hence, we can always map the UV-domain back to the original shape \( \mathcal{M} \) easily. 
Therefore, \( (\pi,\alpha) \) is an equivalent representation to \( \mathcal{M} \).
By rasterizing \( (\pi,\alpha) \) to an image \( {O} \in \mathbb{R}^{R \times R \times 4} \), where \( {O}[i,j] = (\pi[i,j], \alpha[i,j]) \), we can approximate \( \mathcal{M} \) with \( \mathcal{M}^* \), where \( \mathcal{M}^* \) is the triangular mesh reconstructed from \( {O} \) via \emph{remeshing}, where we connect $\pi[i,j], \pi[i,j+1], \pi[i+1,j]$ and \(\pi[i+1,j+1], \pi[i,j+1], \pi[i+1,j]\) to form triangles if the occupancy of the triplet is all $1$. In theory, as \( R \to \infty \), \( \mathcal{M}^* \) will be infinitely close to \( \mathcal{M} \).
This forms the geometry part of the \textit{omage} representation. The material part of an \textit{omage} consists of albedo (3 channels), normal (3 channels), metalness (1 channel), and roughness (1 channel) maps. Together, we obtain a 12 channel omage ${O}^*$ which can be meshed back to a photo-realistic 3D object, as shown in ~\cref{fig:method}.

With the 3D objects encoded as omages, we aim to train an image diffusion model to model the distribution of the 3D objects.  In the next subsections, we will first discuss our design choice for the generative model, and then show how the omages are obtained.

\subsection{Generative modeling for omages}
\label{sec:method-generation}
We observe that generating Object Images (omages) combines aspects of ordinary image generation and set generation.
Within each patch, the generation process resembles standard image generation due to regular connectivity.
However, among the patches, the problem behaves more like set generation: the patch's location in 2D does not affect the 3D shape.
Patches can be swapped and moved around without altering the 3D geometry.
Additionally, touching boundaries in 3D between two patches often sit far apart in 2D, requiring long-range dependency modeling.
Since transformers excel at learning sets and modeling long-range dependencies, and diffusion models are well-known for their image generation capabilities, we use the Diffusion Transformer~\cite{Peebles2022DiT} as our architecture.
Unlike the original method, we set the patch size to 1 to avoid jagged edges in the generation results.

Given the importance of geometry in omages, we first train a model to generate the four geometric channels. We then train a second model to generate the remaining eight channels.
In the second stage, the input has 12 channels, using the first four channels as conditions and excluding them from noise addition and loss computation.

\subsection{Obtaining object images}
\label{sec:method-object-images}
3D objects with UV maps cannot be directly converted into images due to issues such as overlapping regions, out-of-boundary UVs, touching boundaries, or excessive patches. To address this, we use a UV-atlas repacking method with special care to pack patches with material maps into a $(1024,1024,12)$ omage. To avoid large number of patches, we merge vertices with the same 3D and 2D UV coordinates, and keep a maximum of $K$ largest patches.  Detailed descriptions of this process are provided in~\cref{sec:supp_repacking} of the supplement. For efficient learning, we downsample the images with sparse pooling, which snaps the boundaries and eliminates gaps. Further details are provided below.

\begin{figure*}[t]
\centering
\includegraphics[trim={0 0 0 20px},clip,width=1.0\linewidth]{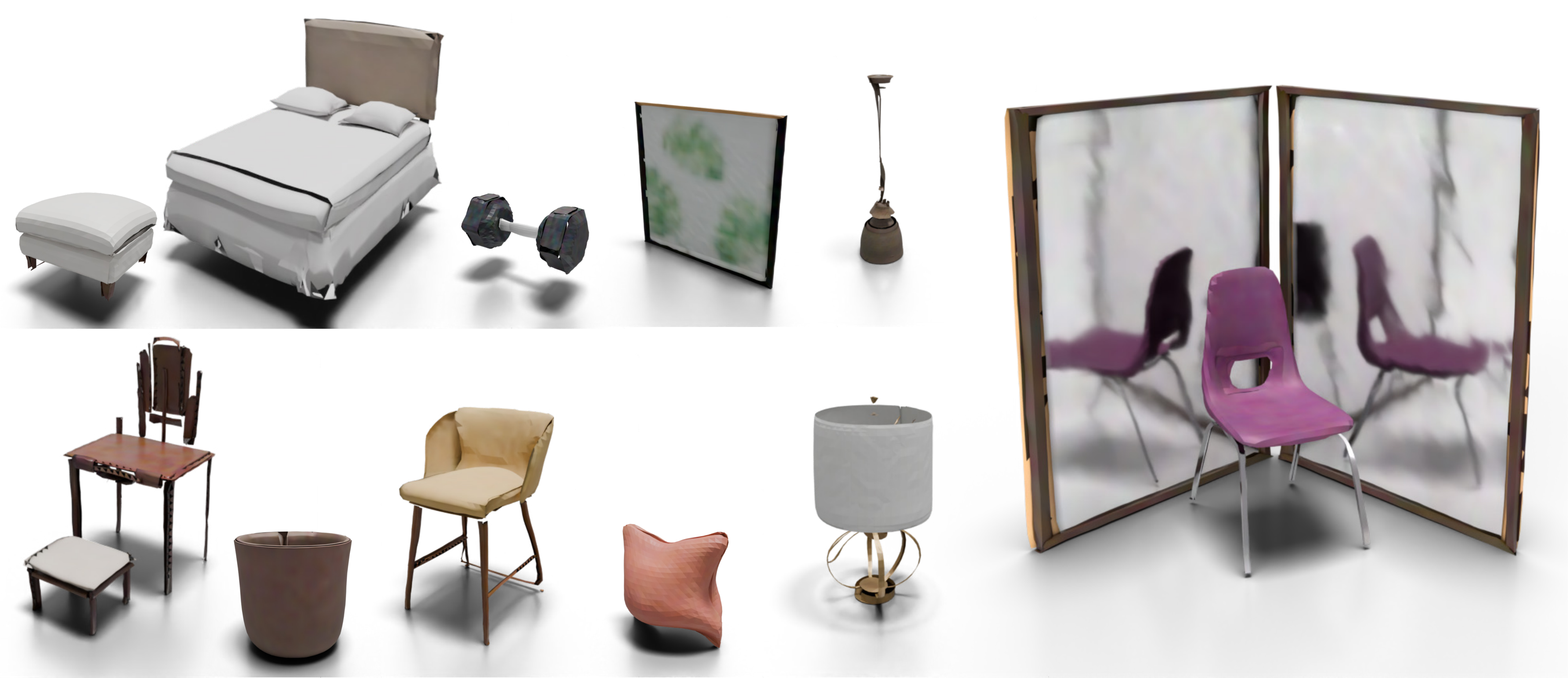}
\vspace{-15pt}
\caption{Examples of label-conditioned Omage-64 generation results. The left side displays results for `ottoman', `bed', `exercise equipment', `painting', `lamp'
`vanity', `plant pot', `chair', `pillow' and `lamp'. 
Even at this resolution, thin structures are successfully generated. On the right, a scene with three objects generated by our method is shown, highlighting our capability in material generation.
}
\label{fig:gallery}
\end{figure*}

\paragraph{Downsample object images and boundary snapping}
Operating within the image domain offers the intrinsic benefit of multi-resolution support. By simply rescaling the omage, object resolution can be adjusted accordingly. For training, we downscale high-resolution omages from 1024 to 64 pixels, enabling efficient processing by transformer models.
As illustrated in \cref{fig:edgesnapping}, standard rescaling methods often fail to preserve boundary information, leading to notable gaps between patches. Inspired by MCGIM~\cite{sander2003multi}, we address this challenge through boundary snapping, where boundary pixels are adjusted based on the contours of the high-resolution image. While this approach is less accurate than using the ground truth mesh boundaries as MCGIM does, it offers greater convenience.
Assuming the higher resolution is divisible by the lower, each pixel in the low-resolution image corresponds to a block of pixels. 
In our case, the block is 16x16. We determine the value of each pixel in the lower resolution image via sparse pooling, averaging only the boundary pixels within each 16x16 block while ignoring other values. This process is illustrated in \cref{fig:edgesnapping} (f) and (g).

\begin{figure*}[t]
\centering
\includegraphics[trim={0 70px 0 8px},clip,width=1.0\linewidth]{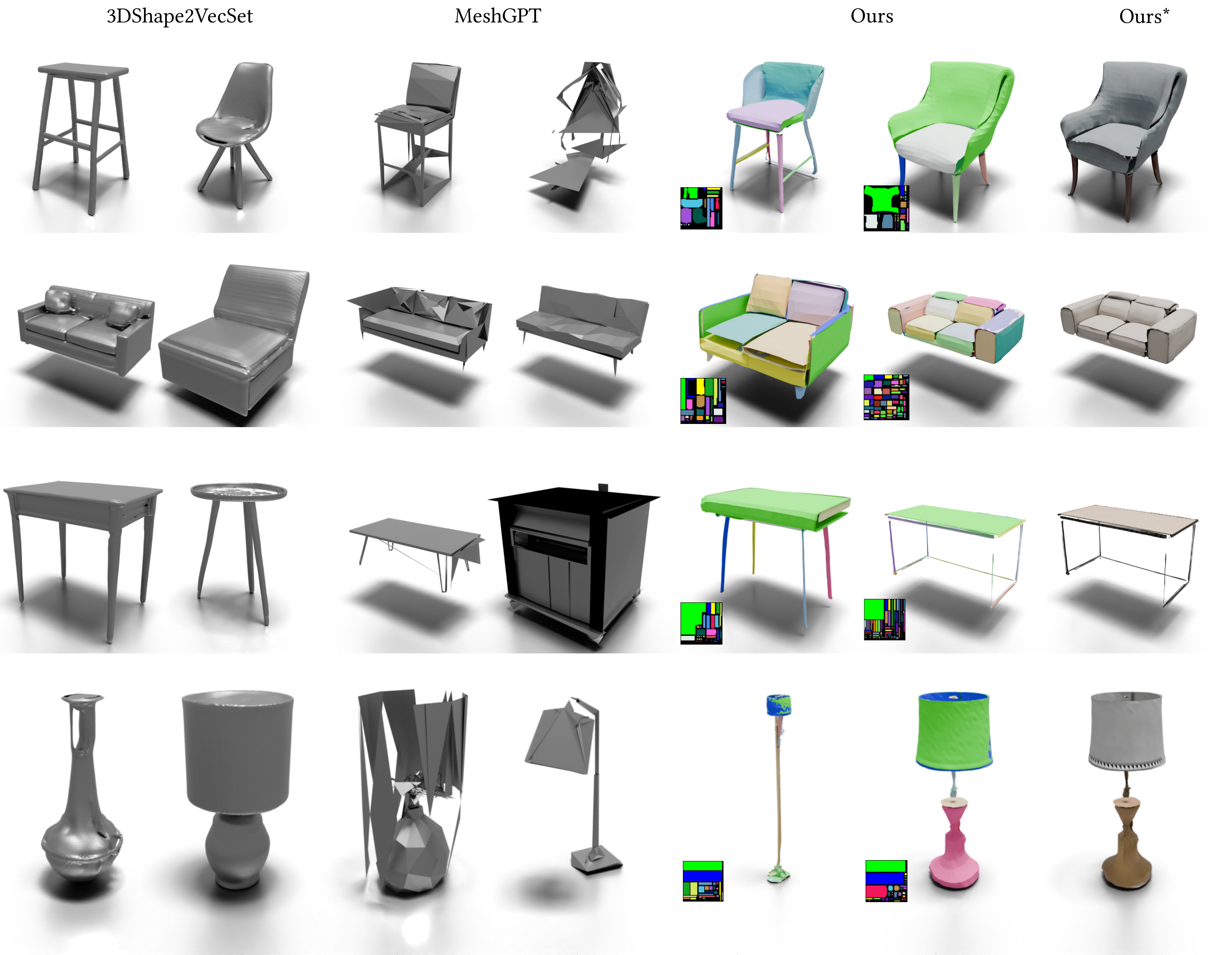}
\vspace{-15pt}
\caption{Label conditioned generation results for \emph{chair}, \emph{sofa}, \emph{table}, and \emph{lamp}.  For our method, we show generated patches in different colors, and with generated material.  Using Object Images, we are able to generate fine detailed geometry with material information.  In contrast, MeshGPT~\cite{siddiqui2023meshgpt} often fails to generate coherent geometry.  3DShape2VecSet~\cite{zhang20233dshape2vecset} generates cleaner geometry but is not able to generate material and patch decomposition.
}
\label{fig:categen}
\end{figure*}

\begin{table*}[ht]
\centering
\caption{Evaluation on class conditional generation.  We measure the point cloud FID (p-FID) and KID (p-KID) for uncolored points sampled from the generated mesh. In geometry generation, with 64 resolution images, we outperform MeshGPT (mGPT)~\cite{siddiqui2023meshgpt} and slightly underperform 3DShape2VecSec (S2VS)~\cite{zhang20233dshape2vecset}.}
\label{tab:categen}
\small

\begin{minipage}{\linewidth}
\begin{adjustbox}{width=\linewidth,center}
    \begin{tabularx}{\linewidth}{@{}l*{15}{X}@{}} %
    \toprule
    & \multicolumn{3}{c}{Chair} & \multicolumn{3}{c}{Sofa} & \multicolumn{3}{c}{Table} & \multicolumn{3}{c}{Lamp} & \multicolumn{3}{c}{Mean} \\
    \cmidrule(lr){2-4} \cmidrule(lr){5-7} \cmidrule(lr){8-10} \cmidrule(lr){11-13} \cmidrule(lr){14-16}
    & S2VS & mGPT & Ours & S2VS & mGPT & Ours & S2VS & mGPT & Ours & S2VS & mGPT & Ours & S2VS & mGPT & Ours \\
    \midrule
    p-FID~$\downarrow$ & 15.9 & 31.2 & 18.9 & 20.6 & 24.9 & 23.3 & 11.9 & 20.3 & 22.4 & 33.0 & 51.2 & 43.6 & 20.4 & 31.9 & 27.0 \\
    p-KID~$\downarrow$ & 7.31 & 17.3 & 7.83 & 9.22 & 10.7 & 9.69 & 2.43 & 7.12 & 6.75 & 14.3 & 31.4 & 26.4 & 8.32 & 16.6 & 12.7 \\
    \bottomrule
    \end{tabularx}
\end{adjustbox}
\end{minipage}
\end{table*}

\begin{figure*}[t]
\centering
\includegraphics[trim={0 0px 0 0px},clip,width=1.0\linewidth]{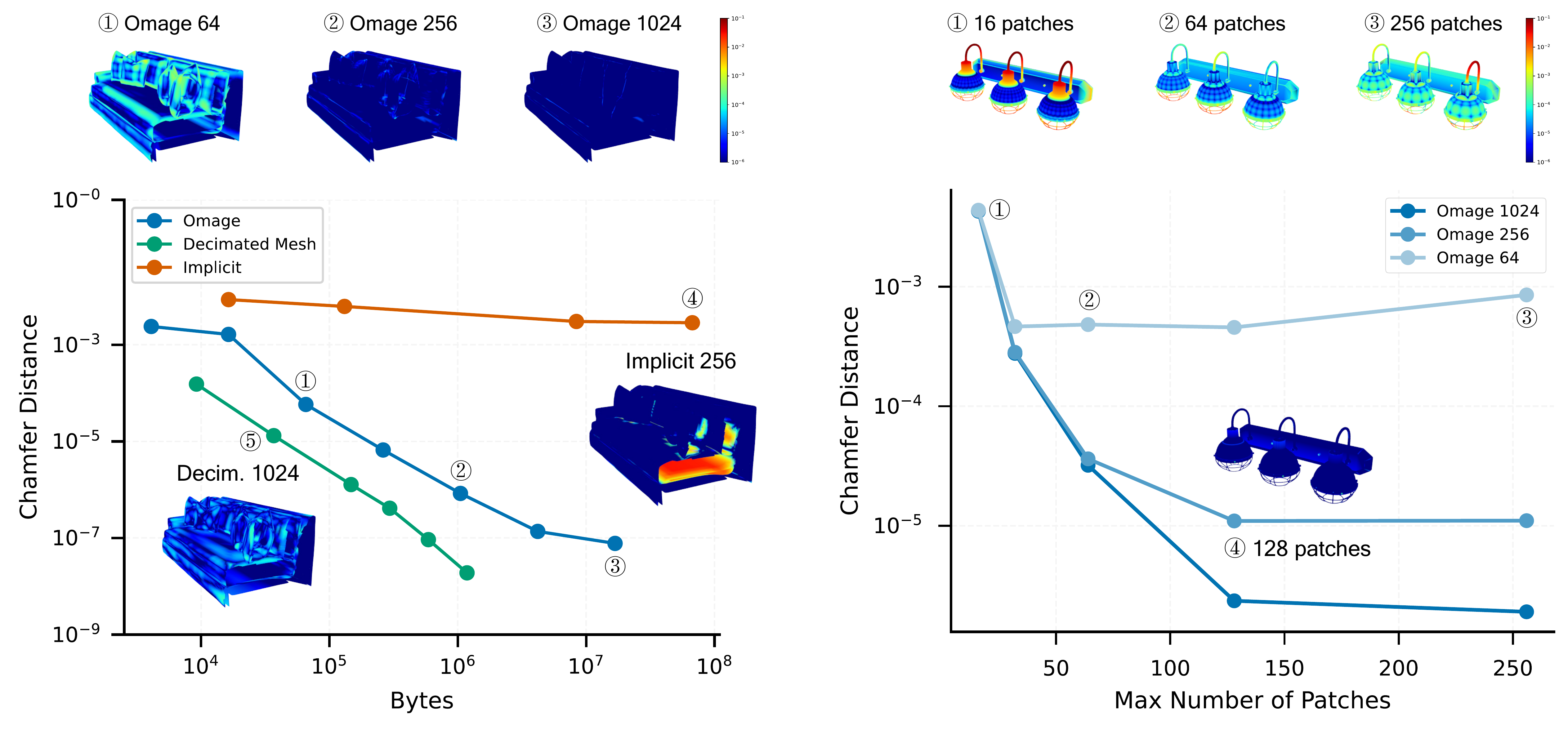}
\vspace{-15pt}
\caption{Representation analysis. Left: Chamfer Distance (CD) vs. byte size. A sectional view of the sofa example highlights the accuracy for both exterior and interior structures. Right: The effect of the maximum number of patches on the accuracy of Omage representations, demonstrating the trade-offs of this parameter. Note that the color map is displayed in log-scale.
}
\label{fig:repr_analysis}
\end{figure*}

\begin{figure}[t]
\centering
\includegraphics[width=1.0\linewidth]{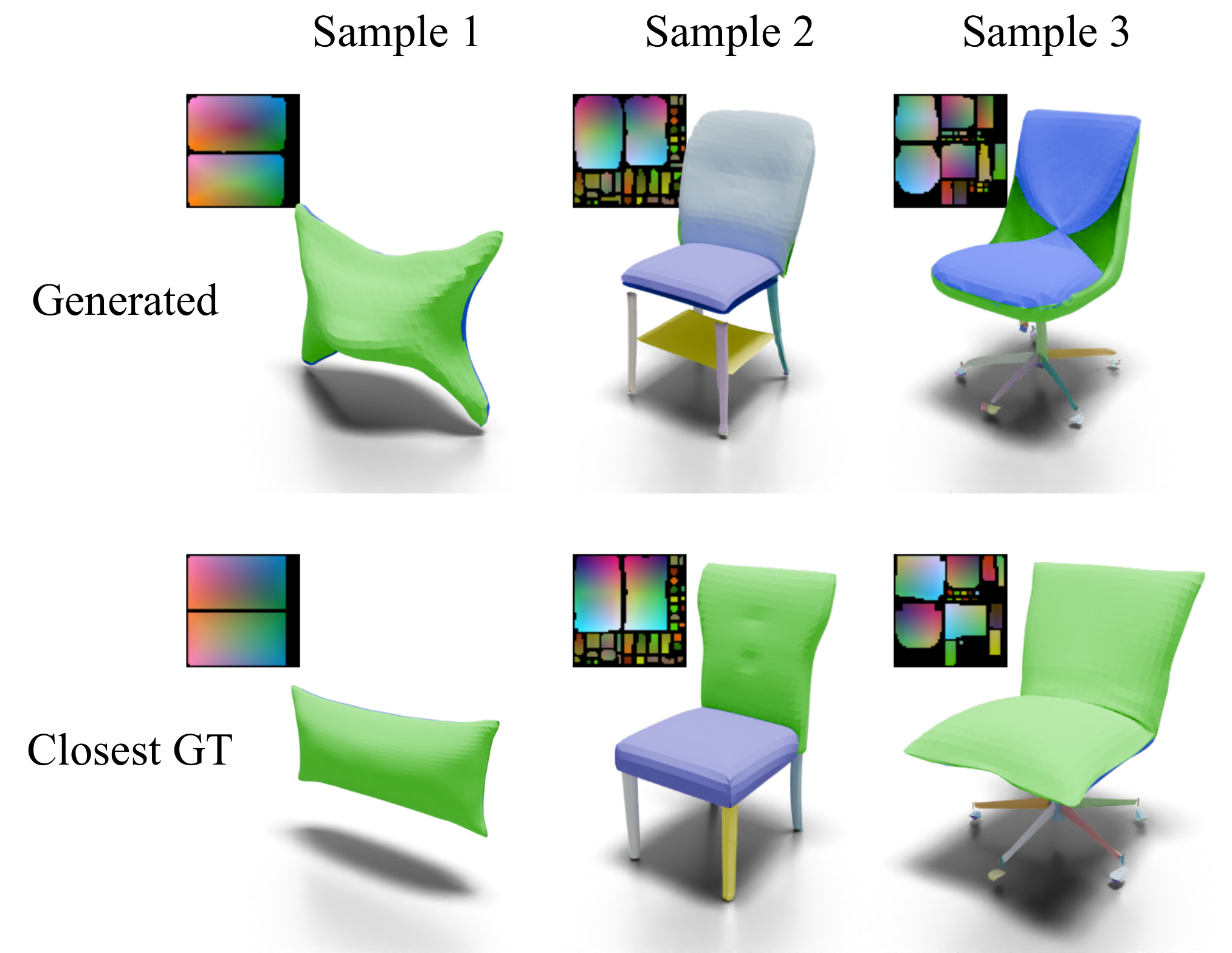}
\vspace{-15pt}
\caption{Our generated results compared with its nearest neighbour in the dataset.}
\label{fig:fig_shapenovelty}
\end{figure}

\section{Experiments}
\label{sec:experiments}

\subsection{Implementation Details}
\label{sec:expr-setup}
\paragraph{Dataset}
We conduct experiments on the Amazon Berkeley Objects (ABO)~\cite{collins2022abo} dataset (license CC BY 4.0) which consists of roughly 8000 high-quality designer-made 3D models with UV-atlases across 63 categories. All of these objects are textured meshes accompanied with initial unprocessed UV-atlases and PBR materials. We convert the glb format shapes to 12 channel 1024 resolution omages with Blender 4.0\footnote{\url{https://www.blender.org/}} using the method described in~\cref{sec:method-object-images}. The 1024 omages are downsampled to 64 resolution with edge snapping. Unlike volumetric representations, the processing of omages is highly efficient and robust. We can obtain 1024 omage from a single raw glb file within 6 seconds. 

\paragraph{Diffusion Transformer architecture and training}
We use DiT-B/1~\cite{Peebles2022DiT} model which has 12 layers of Transformer blocks. We set the patch size to 1 to avoid results with jaggies. This essentially removes the patchify layer, resulting in a full 4096 sequence length. With the help of mixed-16 bit precision training, we train our model with 4 NVIDIA 3090 GPUs for 3 days. We use AdamW~\cite{loshchilov2017decoupled} optimizer with learning rate set to 1e-4. The effective batch size is 32. For generation, we use a classifier-free guidance scale of 4, and 250 sampling steps.

\subsection{\textbf{Class conditional generation}}
\label{sec:expr-class-generation}
In~\cref{fig:gallery}, we present generated results from our model trained on all categories of the dataset. The geometry and material are generated in an autoregressive manner.  With a single representation, our method is able to generate challenging materials such as mirrors (see \cref{fig:gallery} right). 
For evaluation and comparison, we focus on a subset of the four largest categories ('chair', 'sofa', 'table', and 'lamp'), comprising approximately 3800 shapes. We train both our method and the baseline methods on this subset.

\paragraph{Evaluation metrics}
Following previous works~\cite{nichol2022pointe,zhang20233dshape2vecset,yariv2023mosaic}, We use point cloud FID (p-FID) and KID (p-KID) to measure the quality of the generation results. We adopt the pretrained PointNet++~\cite{qi2017pointnet++} feature extractor provided by Point-E~\cite{nichol2022pointe} for calculating FID and KID.
We randomly generate 512 shapes using each model and calculate the metrics for these 512 shapes versus the training set of the categories.

\paragraph{Baselines}
We compare to 3DShape2VecSet~\cite{zhang20233dshape2vecset}, which is one of the state-of-the-art neural implicit-based 3D generative models. Its representation module encodes a 3D occupancy field into a set of latent vectors. 
We also compare to MeshGPT~\cite{siddiqui2023meshgpt}, which uses graph convolutional autoencoder to turn triangle mesh generation into a sequence generation problem.
We refer to our model for comparison as `omage64-DiT'.

For 3DShape2VecSet, we adopt the official implementation from the authors. More specifically, we directly use their autoencoder without additional training and finetune their pretrained diffusion model on ABO dataset. For MeshGPT, we use a third-party implementation\footnote{\url{https://github.com/MarcusLoppe/meshgpt-pytorch}}, which can be trained on shapes decimated to 400 faces. We finetune both its autoencoder and auto-regressive transformer on the ABO dataset. 

As shown in~\cref{fig:categen}, 3DShape2VecSet can generate good quality shapes, but may fail to generate reasonable thin structures (the lamp's wire). Also, the generated shape has very dense triangles due to its implicit nature. Meanwhile, MeshGPT can obtain very compact results (table and sofa), but is prone to have messy triangles. MeshGPT may also generate flipped triangles (see the second table). In contrast, our method can directly generate thin structures and open surfaces.
\Cref{tab:categen} shows that despite the difficulty of structured geometry generation, our method can still achieve similar p-FID and p-KID scores to 3DShape2VecSet.  In addition, our method generates realistic PBR materials and semantically meaningful patch decomposition.

\subsection{Shape Novelty}
In~\cref{fig:fig_shapenovelty}, we check if our method can generate novel samples by comparing generated examples with their closest ground-truth examples in the dataset.
We retrieve the nearest neighbour by directly computing the mean square errors between the generated omage and the omages in the dataset. Our generated result has non-trivial differences toward the nearest neighbours in the training set, showing that our method is not overfitting. However, maybe due to the challenge of the combinatorial nature of omages, the layout of the 2D patches is similar.
The third sample of~\cref{fig:fig_shapenovelty} can be regarded as a failure case of our method: It occasionally connects wrong side of the patch boundary. However, this example still demonstrates an interesting patch alignment.

\subsection{Representation Analysis}
We also analyze the the ability of our representation to capture details of the shape geometry at different resolutions.  
The left side of~\cref{fig:repr_analysis} illustrates a key drawback of implicit representation: it fails to differentiate touching parts, considering all nearby regions as inside, thus failing to reconstruct surfaces like the cushion of a sofa resting on the seat. In contrast, omage preserves such structures effectively and is more efficient, achieving comparable performance to decimated triangle mesh.
On the right side of the figure, we show that the maximum number of patches is also a critical parameter. If too low, large patches are removed, causing significant errors. If too high, particularly for complex shapes with many intricate parts, the gap ratio increases, and pixel density per patch decreases, leading to reduced accuracy in those regions. We choose $K=64$ for 64-resolution omages generation since it strikes a good balance between patch coverage and per-patch accuracy.

\section{Conclusion}
\label{sec:conclusion}
In this paper, we introduced a new paradigm for generating photo-realistic 3D objects with patch structures. We show the possibility of generating 3D object with materials by only denoising a small 64x64 2D image with an image diffusion model.
This new paradigm also has limitations:
It can not guarantee to generate watertight meshes, requires 3D shapes for training to have good quality UV atlases, and the current resolution is only limited to 64. In the future, we will continue to explore how to address these problems to fully utilize the benefits of this regular representation for high-quality structured 3D assets generation.

\ifcase\papermode
\or
    \paragraph{Acknowledgements} We thank Xueqi Ma and Biao Zhang for their advice and guidance in training the baseline methods. This work was supported by a CIFAR AI Chair, an NSERC Discovery grant, and a CFI/BCKDF JELF grant. \textit{Mesh credits}: \cref{fig:repr_cmp}~Headphone~\cite{Kantarci2018headphone}.
\or
    \paragraph{Acknowledgements} We thank Xueqi Ma and Biao Zhang for their advice and guidance in training the baseline methods. This work was supported by a CIFAR AI Chair, an NSERC Discovery grant, and a CFI/BCKDF JELF grant. \textit{Mesh credits}: \cref{fig:repr_cmp}~Headphone~\cite{Kantarci2018headphone}.
\fi

{
    \small
    \bibliographystyle{h_ieeenat_fullname}
    \bibliography{main}
}

\clearpage
\setcounter{page}{1}

\twocolumn[{%
\renewcommand\twocolumn[1][]{#1}%
\maketitlesupplementary

\appendix

\captionsetup{type=figure}\includegraphics[width=\linewidth]{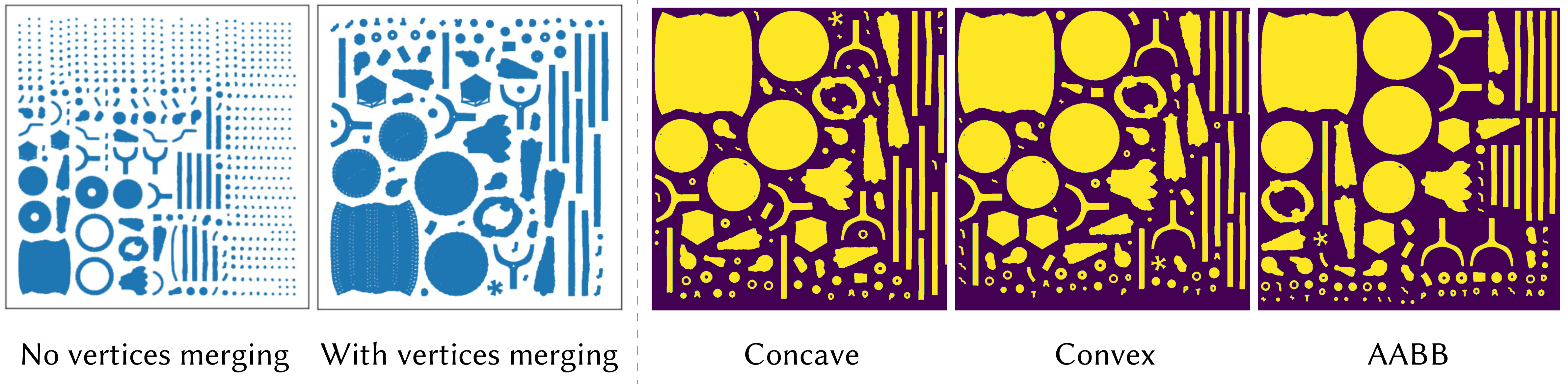}
\vspace{-1em}
\captionof{figure}{\textbf{Left}: Merging coincident vertices before repacking will substantially reduce the number of patches. \textbf{Right}: The results of three commonly used uv-islands packing algorithms.  For our Object Images, we use AABB with vertices merging.
\vspace{.2cm} %
}
\label{fig:repacking}

\captionsetup{type=figure}\includegraphics[width=\linewidth]{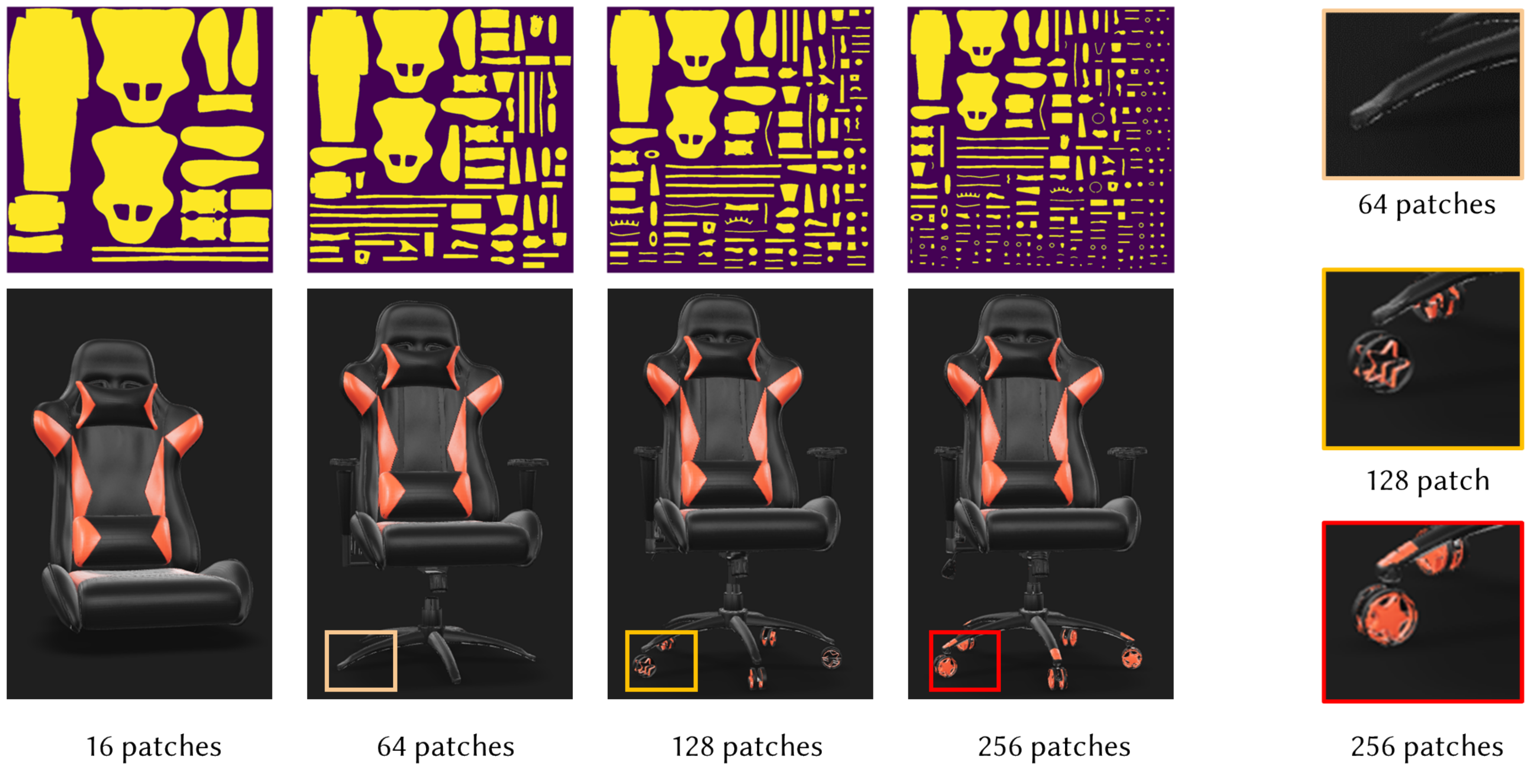}
\vspace{-2em}
\captionof{figure}{The effect of number of kept patches. As the number of patches goes up, more intricate geometric parts are kept. However, the average number of pixels dedicated to each part is reduced. \vspace{.2cm} %
}
\label{fig:numpatch}

}]

\section{Repacking the UV-atlas}
\label{sec:supp_repacking}
As mentioned in~\cref{sec:method-object-images} of the main text, 3D objects with UV-maps generally cannot be directly converted into Object Images (omages) due to issues such as overlapping regions, out-of-boundary UVs, touching boundaries, or extremely large number of patches. Overlapping regions breaks the single-valued assumption, making it impossible to map the overlapped region back to the 3D surface. Since designers often reuse textures for similar parts, overlapping UV islands are common in 3D assets.

Another common issue is the touching boundary problem. One important assumption of omages is that different patches not only do not overlap but can also be separately recognized. We detect different parts by identifying the connected components within the alpha (occupancy) map. If two patches have touching boundaries, this detection will fail, introducing artifacts that connect patches which could be far apart.  To address the above issues, we leverage standard UV-atlas repacking to obtain non-overlapping patches and pack them into high-resolution images.  For efficient learning, we then downsample the images using sparse pooling to snap the boundaries and eliminate gaps.  We describe the repacking and baking step in detail below (the downsampling is described in the main paper).   

\paragraph{Repacking and baking}
We use UV-atlas repacking to obtain clean patches that are free from artifacts.
We first obtain the 2D UV-islands of all patches, then use a 2D irregular shape packing algorithm to rescale and rearrange the UV-atlas within the standard UV-domain, leaving margins between each island. In \cref{fig:repacking} (right), we show the the three packing methods provided by Blender: Concave, Convex, and AABB. Their names indicate the shapes approximations used for the packing of the patches and result in different space utilization efficiency. Concave (exact shape) has the least empty space but introduces complex combinatorial patterns that are challenging for generative models to learn. Hence, we adopt AABB as the primary method for repacking.

Another common issue is that many patches are unnecessarily separated into multiple sub-patches by default. This results in numerous small pieces that degrade the quality of the omage, potentially reducing it to a triangle soup or point cloud as the number of patches increases. By merging vertices that share the same 3D and 2D UV coordinates, we can reconnect these sub-patches to form larger patches. This not only reduces empty space but also improves the integrity of the patches.  See \cref{fig:repacking} (left) for comparison of packing with and without vertex merging.

After merging the sub-patches, there may still be an excessive number of patches. To simplify the generative modeling, we keep a maximum number of patches, $K$. For shapes with more patches than this threshold, we sort the patches by their 3D area and retain only the largest $K$ patches. \cref{fig:numpatch} shows the effect of this parameter. Having more patches preserves geometric details but complicates generative modeling. This is especially true for lower-resolution omages, where smaller parts lack enough pixels to form meaningful regions. In practice, we find that a maximum of 64 patches works well for generating 64-resolution omages (See~\cref{fig:repr_analysis}).

After repacking, we rasterize the geometry and material properties into an image format through texture baking according to the repacked UV-atlas. We bake the geometry (including UV occupancy), normal map, albedo, metalness, and roughness into the final $(R, R, 12)$ omage, with $R=1024$ set as default for high-quality results.

\end{document}